\pdfoutput=1
\documentclass{article}
\usepackage[numbers]{natbib}

\usepackage[preprint]{neurips_2019}

\usepackage[utf8]{inputenc} 
\usepackage[T1]{fontenc}    
\usepackage{hyperref}       
\usepackage{url}            
\usepackage{booktabs}       
\usepackage{amsfonts}       
\usepackage{nicefrac}       
\usepackage{microtype}      
\usepackage{makecell}
\usepackage{easytable}
\usepackage{graphicx}
\usepackage{amsmath,amssymb}

\DeclareMathOperator{\EX}{\mathbb{E}}

\title{Grayscale Image Colorization with GAN and CycleGAN in Different Image Domains}

\author{%
        Chen~Liang\\
        Carnegie Mellon University\\
        Pittsburgh, PA 15213 \\
        \texttt{chenlia2@andrew.cmu.edu}
 \\
 \And 
        Yunchen~Sheng\\
        Carnegie Mellon University\\
        Pittsburgh, PA 15213 \\
        \texttt{yunchens@andrew.cmu.edu}
  \\
 \And 
        \thanks{Acknowledgement: Special thanks to Geyang Zhang for her generous help} 
        Yichen~Mo\\
        Carnegie Mellon University\\
        Pittsburgh, PA 15213 \\
        \texttt{yichenmo@andrew.cmu.edu}}

\begin{document}

\maketitle

\begin{abstract}
Automatic colorization of grayscale image has been a challenging task. Previous research have applied supervised methods in conquering this problem ~\cite{Koo2016}. In this paper, we reproduces a GAN-based coloring model, and experiments one of its variant. We also proposed a CycleGAN based model and experiments those methods on various datasets. The result shows that the proposed CycleGAN model does well in human-face coloring and comic coloring, but lack the ability to diverse colorization. 
\end{abstract}

\section{Introduction}
Grayscale image colorization is assigning colors to the black and white images. Original methods for image colorization mainly involve in human efforts and traditional image processing methods. The development of deep learning enables automated colorization to be an active research area \cite{Nazeri2018}. Previously, approaches are mainly focused on restoring the original color of the images. Though, we are interested in a generative fashion for image colorization, which could be applied for restoring the black and white films, recreating comic books \cite{Guadarrama2019}.

Several approaches are proposed to generate multiple plausible image colorization, including Conditional Random Field (CRF), Generative Adversarial Network(GAN), or modified CNN. In this project, we choose the state of art network proposed in Cao et. al \cite{Cao2017} as our baseline model. 

We try to reproducing the result proposed in the baseline model paper. However, the results are not as good as the paper states. Therefore, we try to use GAN instead of wGAN. Finally, we proposed a conditional CycleGAN model that do much better than the baseline model.

To evaluation the generalization ability of the proposed model, besides the bedroom images that are used in the baseline model, we also experiment the baseline and CycleGAN models on a human face dataset, and experiment the CycleGAN model on a comic dataset. The experiments show that the proposed CycleGAN overperforms the baseline model in human face coloring, and also do well in comic coloring.


\section{Related Work}
\begin{enumerate}
    
\item Unsupervised Diverse Colorization via
Generative Adversarial Networks \cite{Cao2017} \newline
This paper described the baseline model we are using for coloring gray scale pictures with GAN. We would like to further explore its ability of colorization, in the sense of improving its flexibility. 

\item LSUN dataset \cite{Yu2015} \newline
This is the gray scale image dataset we are using to train our model. The images contained in this dataset are 256 x 256 jpg binary data.

\item Generative adversarial nets \cite{Goodfellow2014} \newline
This paper describes the original design and explanation of GAN. This paper introduces how the adversarial objective function connecting true images and generated images, so we have a sense of how to modify the network as to fit the third property to reach our goal.

\item Learning diverse image colorization \cite{Deshpande2017} \newline
The paper talked about how they embedding a low dimension of color space into the generator such that the grey scale image can be colored as multiple colors. Since we would like to customize our generated image, this paper provide a nice background for image colorization and inspired our project idea.. 

\item Automatic Colorization with Deep Convolutional Generative Adversarial Networks \cite{Koo2016, wang2023sentiment, hu2020antvis} \newline
The current model we are using is in an unsupervised fashion. However, if we would like to assign a specific color scheme to our generated image, we may need to have it as a ground truth in the loss function. This paper describes the advantages and disadvantages of using supervised learning in GAN. \newline

\item InkGan: Generative Adversarial Networks for Ink-And-Wash Style Transfer of Photographs \cite{yu2023inkgan}  \newline

\end{enumerate}{}

\section{Data}
We mainly applied three datasets in evaluating our models. For all datasets we are using YUV representation of the color image. When training as a gray scale image, only maintain the Y dimension and reseed the U, V dimension.
\begin{enumerate}
\item 
LSUN bedroom dataset \cite{Yu2015} is a well established color image dataset. Among all 10 scene and 20 object categories, we choose to use the indoor occasion for its relevantly rich color content, comparing to simple outdoor scene composition. This challenge allows our network to fully express it's capacity.
Data size: The entire data set includes 3,033,042 images in the training set, and 300 images in the validation set. The baseline model randomly picked 503,900 images to train the model. The size of original color image is 256 x 256. We resize the input image to 64 x 64.
\item 
Labeled Face in the Wild dataset, which is available from the website \url{http://vis-www.cs.umass.edu/lfw/}. The entire dataset 
includes 13233 images from 5749 people. We explored wGAN and CycleGAN on this dataset. The original image size is 250*250. No image prepossessing is performed. We trained the CycleGAN model on about 6000 randomly picked images and test the model on 500 randomly picked images.
\item 
The comic dataset is JoJo's Bizarre Adventure Part 5: Golden Wind. There are about 1685 comic images with original image size of 1560*1200. Images are resized to 286 * 286 and then randomly cropped to 256*256 for training and testing. We used 1525 images for training and 160 images as testing data. For CycleGAN, the training data contains a grayscale and a colorized version of the same comic, the colorization is done by JoJo's Colored Adventure Team, which is available from the website \url{http://jojocoloredadventure.blogspot.com/2014/08/download-our-latest-releases.html} 
\end{enumerate}{}
\section{Proposed Solution}




\subsection{WGAN - The Baseline Model}
    
We start from the baseline model\cite{Yu2015} which uses a conditional GANs to color the grey-scale image. Baseline code is based on \url{https://github.com/changjianhui/ColorGAN-update} 
 
In this method, the generator is a fully convolutional network with a single layer noise. The generator takes in the grey-scale image, i.e Y-value of the original image, and a noise vector, then output the U-value and V-value of the image out for the YUV color space, combines them with the Y-value of the image to get the complete color image. Then the generated images are feed it into the discriminator.

\begin{figure}
  \centering
  \includegraphics[scale=0.5]{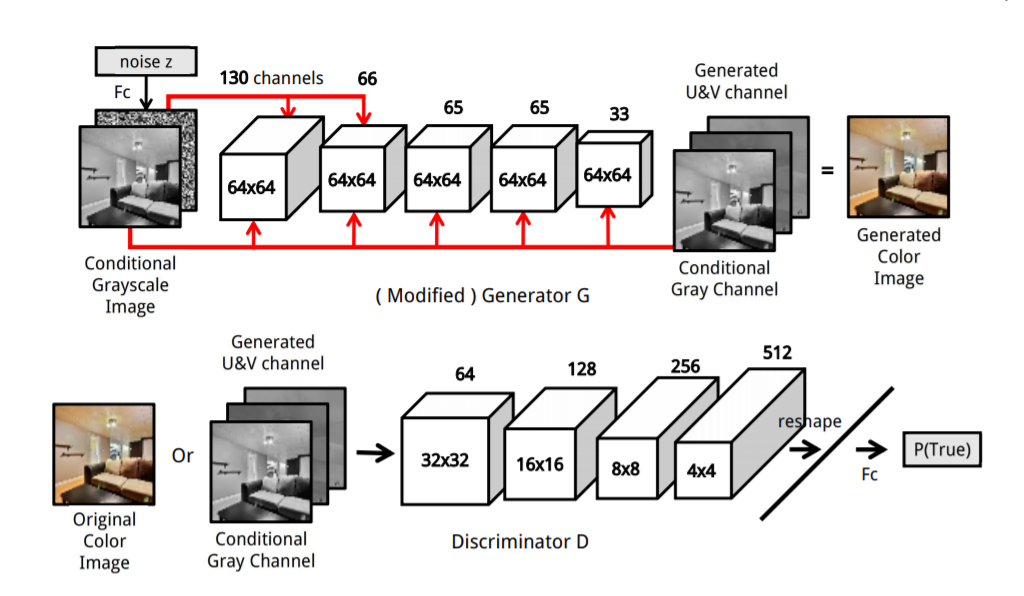}
  \caption{The network structure of the baseline model.}
\end{figure}

The discriminator is trained to distinguish a fake image that is generated by the generator and the ground-truth image with the most real colors. 

The training of both generator and discriminator is performed synchronously, and the baseline model also uses the Wasserstein GAN \cite{Arjovsky2017} to solve the problem of mode collapse and gradient vanishing. With Wasserstein GAN, the loss functions we would like to minimize of the generator and the discriminator are as follows:

\begin{equation}
LOSS_{G} = -D(img_{fake})
\end{equation}
\begin{equation}
LOSS_{D} = -(D(img_{real}) - D(img_{fake}))
\end{equation}

Here G is the generator and D is the discriminator. the output of the discriminator is a scalar logits $D \in \mathbb{R}$, and the larger the value is, the more probable the input image has "real colors". 

\subsection{GAN - Baseline Variant}
Compared with wGAN, GAN tries to minimize a slightly different function with respective to discriminator and generator \cite{Goodfellow2014}. 
\begin{equation}
    \min_G \max_D V(D,G) = \EX_{x \sim p_{data}(x)}[logD(x)] + \EX_{z \sim p_z(z)}[1-log(D(G(z)))]
\end{equation}{}

For original GAN, sigmoid function is applied to the discriminator. Besides, we need to calculate logarithm for the discriminator loss and generator loss. Based on the wGAN baseline model, we mainly changed the loss function from the baseline code.

\subsection{Conditional cycleGAN}
\begin{figure}
  \centering
  \includegraphics[scale=0.35]{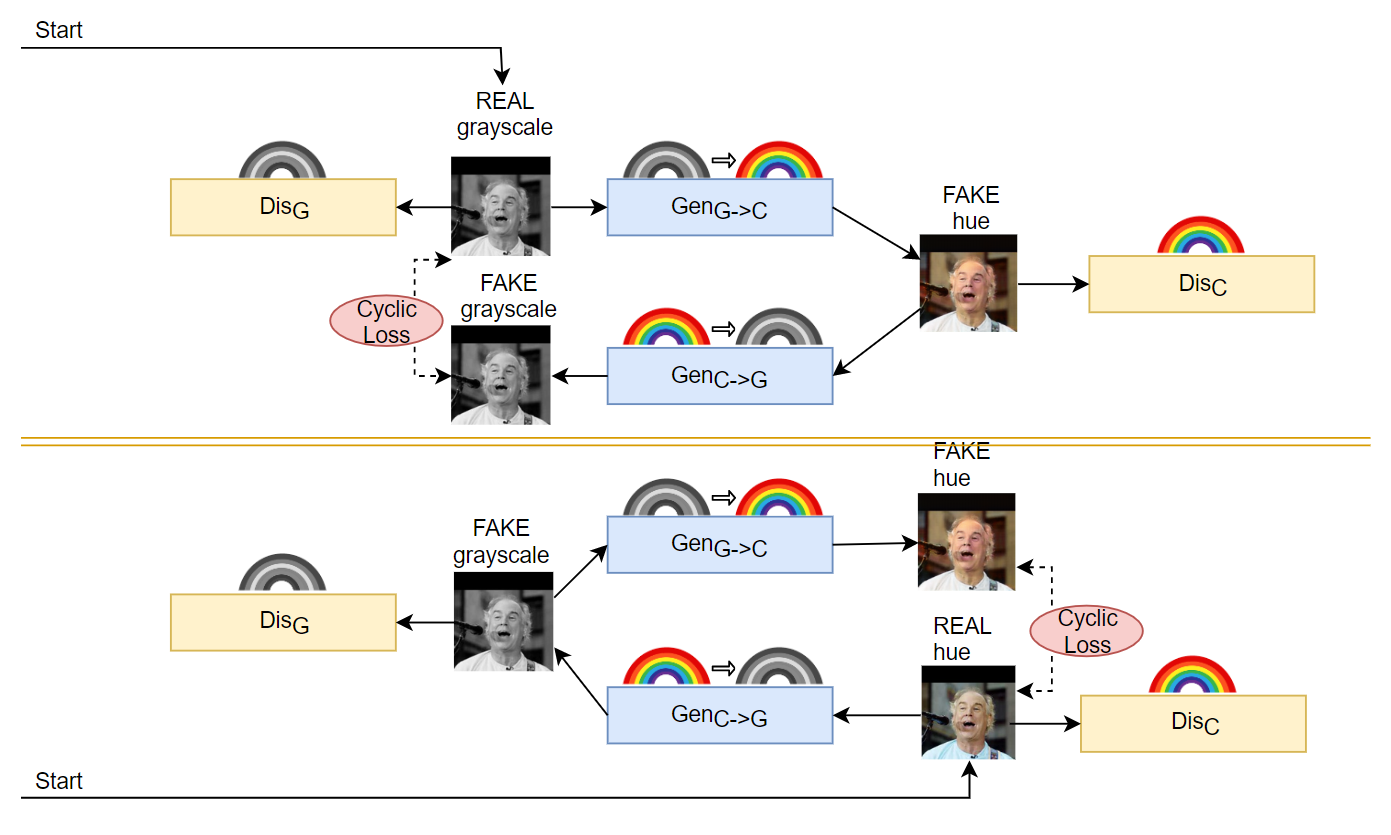}
  \caption{coloring conditional CycleGAN model structure}
\end{figure}

As the above methods don't work well as described in the original paper, we try to use the cycleGAN method and improve the original cycleGAN by adopting the concept of conditional GAN from base line model to improve performance. Original cycleGAN model is initially proposed to do cross-domain image transformation, such as transforming images of horses into images of zebras\cite{CycleGAN}. It is one recent successful approach to do image transformation. 

Generally, CycleGAN transforms images between two domains X, and Y. It is trained to be able to do two transformations: $F: X->Y$ and and $G: Y->X$ while trying to fulfill that: 

(1) For $\forall x \sim p(x)$, $F(x) \sim p(Y)$. For $\forall y \sim p(y)$, $G(y) \sim p(X)$. This means $F$ and $G$ are able to generate convincing images from $X$ to $Y$  and from $Y$ to  $X$ respectively. 

(2) For $\forall x \in X, G(F(x)) = x$. For $\forall y \in Y, G(F(y)) = y$. This means the transformed image of $F$ or $G$ can be transformed back by the other transformation function.

The main difference of cycle GAN and normal GAN is the additional cycle consistency L1 loss, which is calculated as the L1 difference between the true image and generated fake image. Therefore, the losses of cycleGAN are the typical GAN loss combined with the cycle loss:
\begin{equation}
    \mathcal{L}_{GAN}(Gen_{G->C}, Dis_C, G, C)=\EX_{c\sim p(c)}[log(Dis_C(c))]+\EX_{g\sim p(g)}[log(1-Dis_G(g))]
\end{equation}{}
\begin{equation}
    \mathcal{L}_{cycle}(Gen_{G->C}, Gen_{C->G})=\EX_{g\sim p(g)}[|| Gen_{C->G}( Gen_{G->C}(g)) - g||]
    +\EX_{c\sim p(c)}[|| Gen_{G->C}( Gen_{C->G}(c)) - c||]
\end{equation}{}

For our image colorization task, we proposed the "conditonal cycleGAN", which adopt the concept of conditional GAN from base line model. An image with color can be represented in different forms. RGB is the most common representation which splits a pixel into red, green, blue three channels. Alternative representations are Lab and YUV. In this task, we have grayscale image as conditional information, and thus it is straightforward to use YUV. Because the Y and L channel or so called luminance channel which represents exactly the grayscale information. By using YUV, we can just predict UV channels and then concatenate with the grayscale channel to give a full color image.

Specifically, we define these two image domains as $G$: grayscale images  (Y value out of image YUV space) and $C$: images hue(UV value out of image YUV space). Our coloring CycleGAN consists of two generators: $Gen_{G->C}$ and $Gen_{C->G}$, and two discriminators $Dis_{G}$ and $Dis_{C}$. In original CycleGAN, the generator has same number of input and output channel for style transform. However in our method, $Gen_{G->C}$ transforms Y channel(grayscale) to UV channels(hue) while $Gen_{C->G}$ transform UV channels(hue) to Y channel(grayscale). The model structure is in figure 2. Note that a GAN can be extended to a conditional GAN if both the generator and discriminator are conditioned on some extra information. In this case, the extra information is Y channel, and unlike normal GAN, the generator has no random noise (z) input.

Note that there are two GANs in a cycle GAN. In first GAN, the generator is Y to UV and the discriminator is judging real or fake color images using all YUV channels. In the second GAN, the generator is UV to Y, and the discriminator is judging real or fake grayscale images using all YUV channels (which is not necessary, but we keep this structure for symmetric). The training is two GANs in cycle GAN is in parallel. For each input sample, one color image and one grayscale image is picked, and both are split into YUV channels, and they are sent into corresponding generators. The generated fake UV/Y image channel are sent into corresponding discriminators after correct concatenation.

During training, the input image size is set to be 256 by 256 , this is because of memory limit and training time reduction, if we set input size to 720*720, it will have out of memory error even batch size equals to one, and doubling pixel size would result in 4X training time. 

In testing, we only use one generator to generate UV channels from grayscale image's Y channel, the final image is the combination of Y channel and generated UV channels to form YUV color encoding image. 

The conditonal CycleGAN preserves most image information and generator only have to learn the color distribution of image.

\section{Results \& Analysis}

\begin{figure}
  \centering
  \includegraphics[scale=0.4]{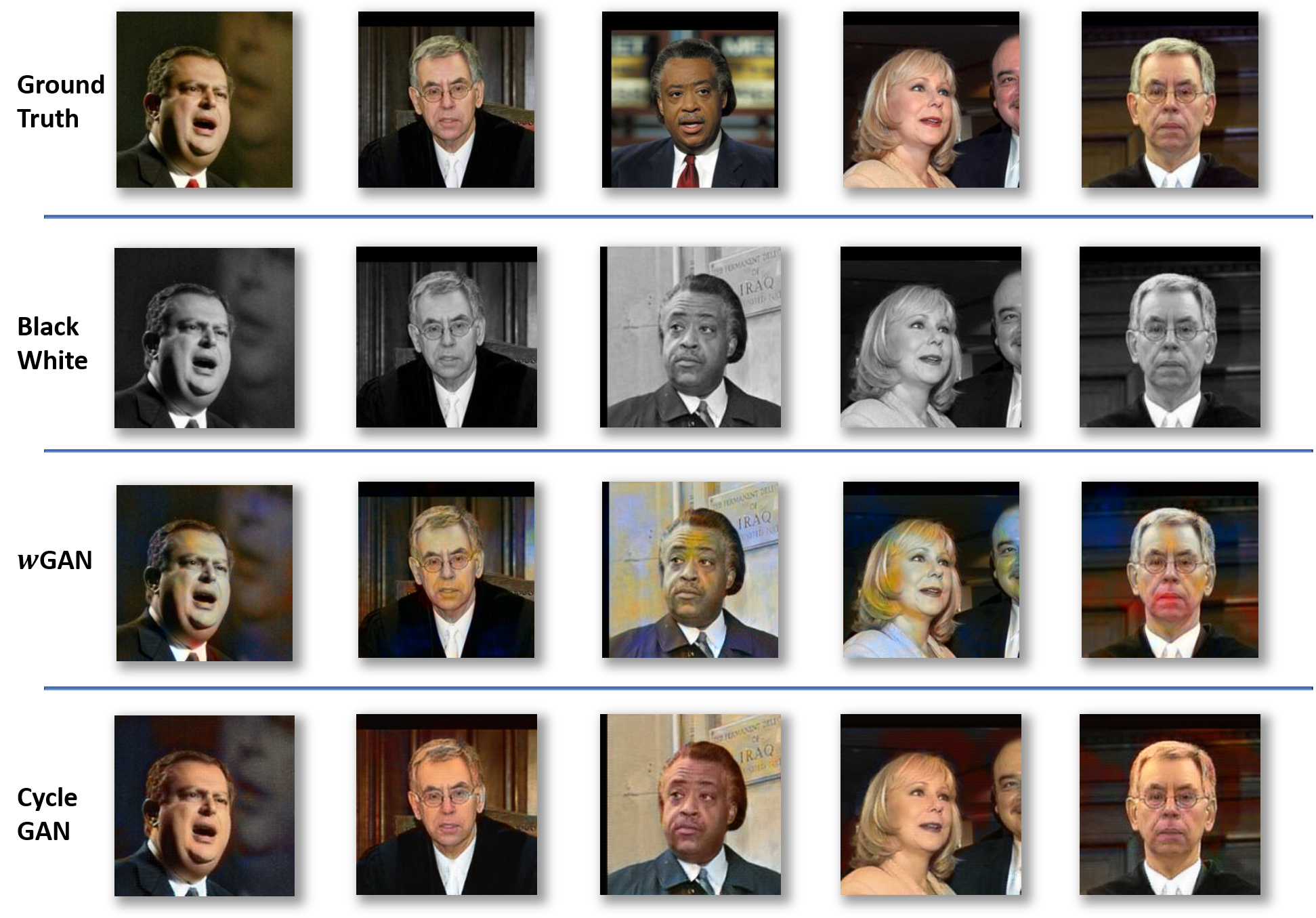}
  \caption{Labeled Face dataset coloring}
\end{figure}

\begin{figure}
  \centering
  \includegraphics[scale=0.5]{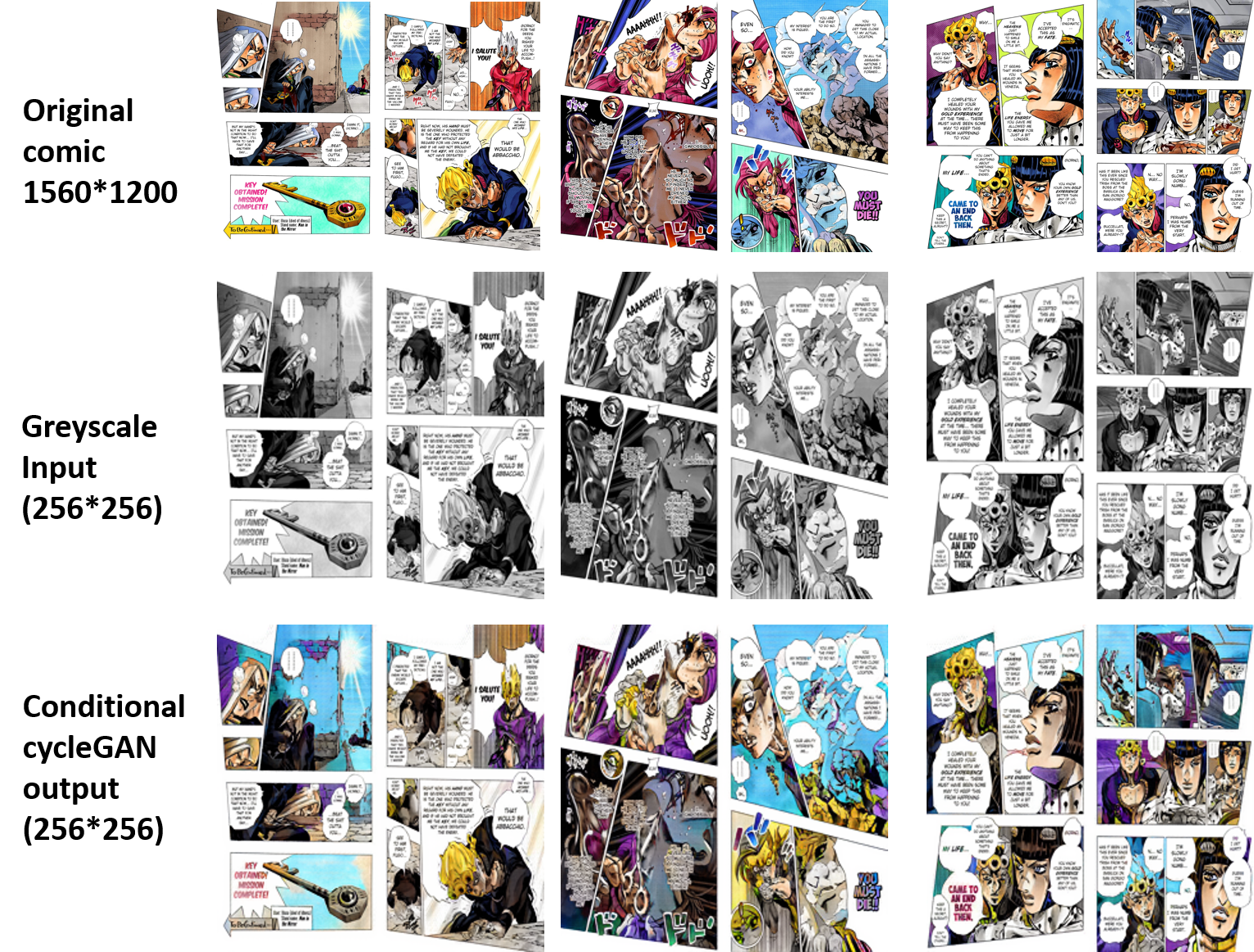}
  \caption{Conditional cycleGAN colorized comic }
\end{figure}

\begin{figure}
  \centering
  \includegraphics[scale=0.5]{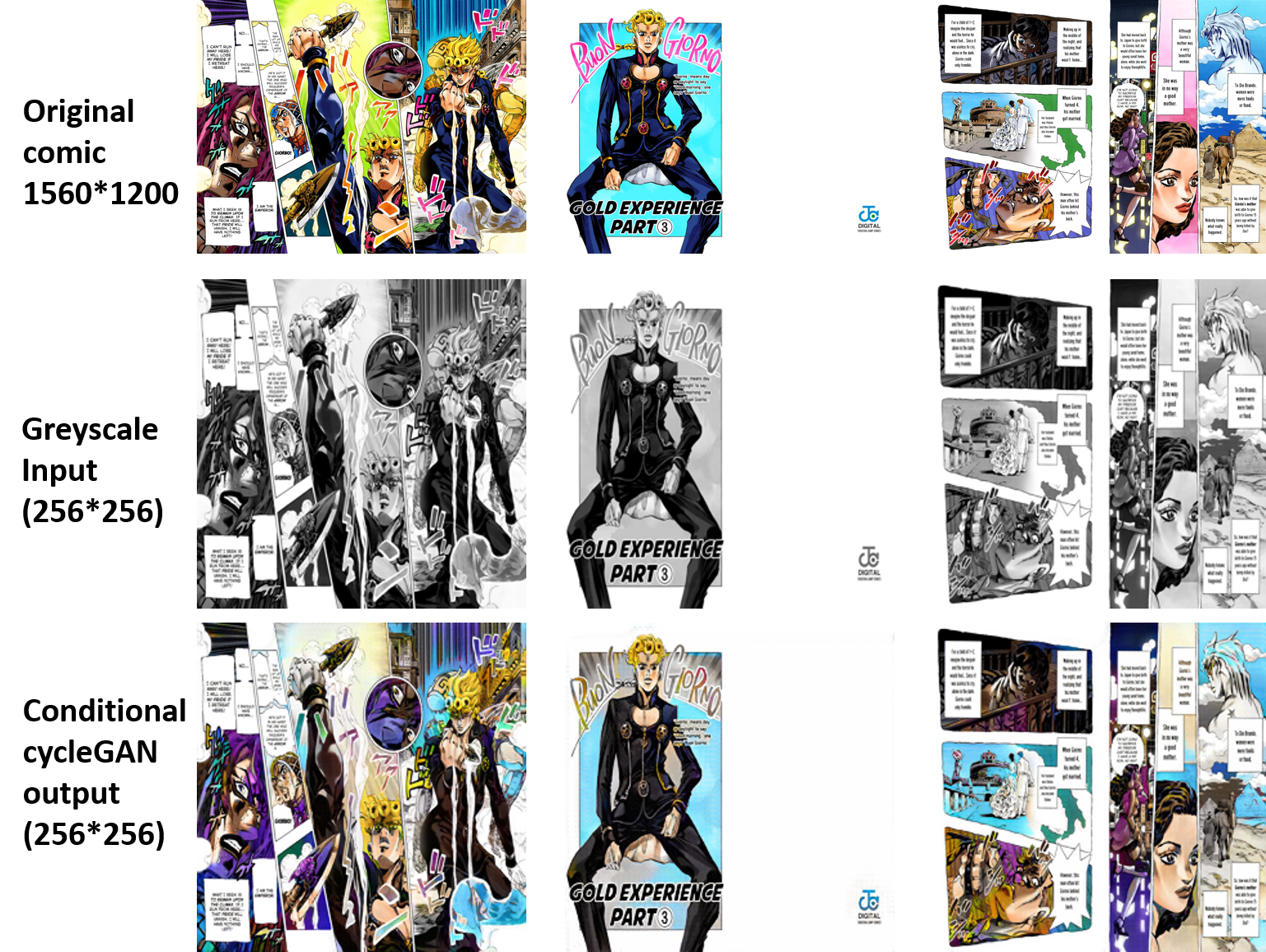}
  \caption{Conditional cycleGAN colorized comic 2}
\end{figure}

\begin{figure}
  \centering
  \includegraphics[scale=0.45]{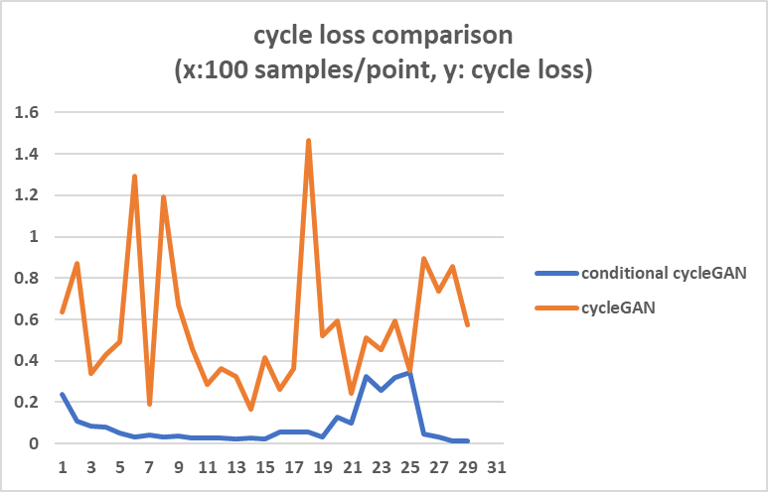}
  \caption{cycle loss comparison}
\end{figure}

\begin{figure}
  \centering
  \includegraphics[scale=0.5]{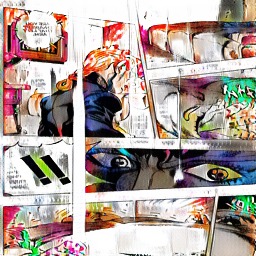}
  \caption{original cycleGAN trying to learn comic image structure instead of color}
\end{figure}
1. Our first experiment is train model on the LSUN bedroom dataset. We try both wGAN method and
traditional GAN to train our baseline model, but the result is bad because all output image has different color blocks on random position of image,so the result is not shown in report.

2. The GAN colorization does not do well for in-door still-life images, the reason might because there are too many items and different colors in a small 256* 256 low resolution image.We suspect that GAN is more suitable to color human faces, since color of face is simpler and consistent. Therefore, we pick the Labeled Face dataset and on which we experiment with our baseline wGAN model, and also our proposed cycleGAN model, the result is at Figure 3.

3. We also try our proposed cycleGAN on comic colorization, which is at Figure 4 and Figure 5.

4. From the result images, we can see that the CycleGAN performance is better than normal GAN and wGAN. The reason might be that, normal GAN can learn the mapping between input distribution X and target distribution Y, however, a mapping function can map a set of inputs to any random permutation of outputs in target distribution, which means there is no constraint to teach the mapping function to map the a particular input image x in distribution X to particular desired output image y in distribution Y. To add more “guideline” while training, cycle-gan introduce cycle loss, and based on our result, it really improve the performance and the output image's color is more stable.

5. During training, we find that the loss and output image of conditional cycleGAN is more stable compared to original cycleGAN, the loss comparison is shown at Figure 6. We can see that conditional cycleGAN has lower loss and the loss is more stable. The original cycleGAN output image shown at Figure 7. We can see that the model is trying to learn the structure of comic image instead of just color of comic image, the white vertical line is caused by model want to learn the comic frame structure.

6. For comic colorization, our model can somehow learn the hair color of the main character in comic. For example, the hair color of the main character Jotaro can keep the same across different output colorized images. And also, if one character is not or seldom seem in training dataset, there is no way we can tell the hair color or cloth color of that character, so our model can not colorize that character correctly.

\section{Conclusion}
In all, we experiment with GAN-based image coloring model and our proposed conditional CycleGAN image coloring model on three different image domains: the in-door bedroom image dataset, the human face dataset, and a comic dataset. The result shows that the conditional CycleGAN produces more plausible images on human face coloring, and also do well in comic coloring.

\small

\bibliographystyle{unsrt}

\bibliography{main}

\end{document}